\documentclass[runningheads]{llncs}
\usepackage{makeidx}
\usepackage{graphicx}
\usepackage{amsmath,amssymb} % define this before the line numbering.
\usepackage{color}
\usepackage{epsfig}
\usepackage{graphicx}
\usepackage{cuted}
\usepackage{float}
\usepackage{authblk}
\usepackage{xspace} 

\DeclareUnicodeCharacter{FB01}{fi}
\newcommand{\etal}{\textit{et al.}}
\newcommand{\ie}{i.e.\@\xspace}
\newcommand{\R}{\mathbb{R}}
\begin{document}
	\pagestyle{headings}
	\mainmatter

	% Insert your submission number here
	\def\GCPR20SubNumber{45}

	% Replace with your title
	\title{Adversarial Synthesis of Human Pose From Text}

	% DO NOT MODIFY these for the draft version that is used for the
	% review process.
	\titlerunning{Adversarial Synthesis of Human Pose From Text}
	\authorrunning{Zhang, Briq, Tanke, Gall}
	\author{Yifei Zhang\textsuperscript{1,2}
	\and
	Rania Briq\textsuperscript{1}
	\and
	Julian Tanke\textsuperscript{1}
	\and
	Juergen Gall\textsuperscript{1}}
	\institute{Computer Vision Group, University of Bonn\\
	{\tt\small briq,tanke,gall@iai.uni-bonn.de}
	\and
	Bonn-Aachen International Center for Information Technology, RWTH-Aachen University\\
	{\tt\small yifei.zhang@rwth-aachen.de}
	}
	
	\maketitle
	
	\begin{abstract}
		This work focuses on synthesizing human poses from human-level text descriptions. We propose a model that is based on a conditional generative adversarial network. It is designed to generate 2D human poses conditioned on human-written text descriptions. The model is trained and evaluated using the COCO dataset, which consists of images capturing complex everyday scenes with various human poses. We show through qualitative and quantitative results that the model is capable of synthesizing plausible poses matching the given text, indicating that it is possible to generate poses that are consistent with the given semantic features, especially for actions with distinctive poses.
	\end{abstract}
	
	\section{Introduction}
	\label{sec:introduction}
	\begin{figure*}[t]
		\begin{center}
			\includegraphics[width=12cm]{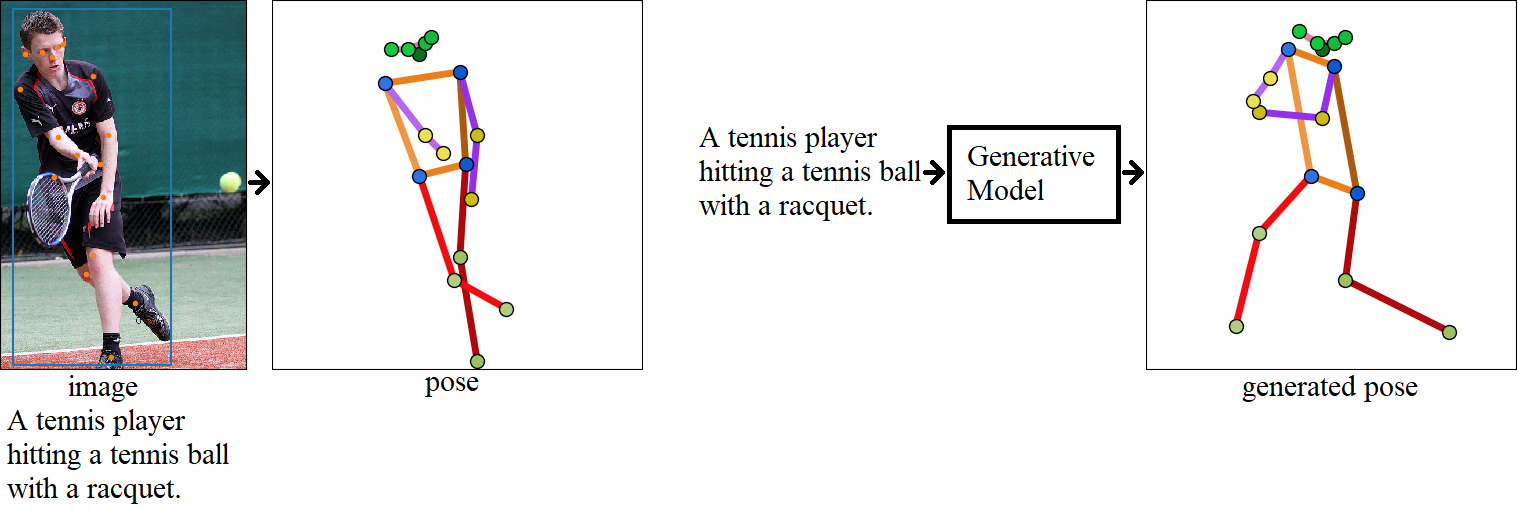}
		\end{center}
		\caption{The image on the left hand side shows an example from the COCO dataset that is annotated by an image caption describing the image and the human pose. In this work, we use only the image caption to generate the human pose.}
		\label{fig:intro_example}
	\end{figure*}
	Given a text description like ``A tennis player hitting a tennis ball with a racquet'', we can directly imagine a human pose that matches the description. Such ability would be useful for applications like retrieving images with semantically similar poses or animating avatars based on text descriptions. Synthesizing the human pose, however, is very difficult since the articulated body pose is much more complex than rigid or nearly convex shapes like objects or faces. Although previous works on synthesizing images from text describing a scene~\cite {Li_2019_story,NIPS2019_8375,qiao2019mirrorgan,reed16,xu2018attngan,zhang17,tan2019semantics} achieve astonishing results when the images contain objects such as flowers, animals with small pose variations like birds or general scenes such as mountains or playing fields, the synthesized humans in these scenes appear quite unrealistic due to distorted or incorrect poses. This failure is due to the uniqueness of the human pose which is highly articulated and versatile. Conversely, most existing works for modeling humans rely on the pose as part of the intermediate feature representation \cite{SMPL:2015,Lassner_2017_ICCV}. Synthesizing poses in complex scenes is therefore an essential step towards synthesizing images with realistic human poses.
	
	In this work, we focus on synthesizing versatile human poses from text as shown in Figure~\ref{fig:intro_example}.  We examine how well the synthesized poses match the text description and whether it is possible to achieve semantic consistency in their feature spaces. To achieve this goal, we design a model based on Generative Adversarial Networks (GANs) \cite{goodfellow14} to generate a single person pose conditioned on a given human-level text description. In order to condition the network to generate a pose that matches the text, the text is first encoded into an embedding using a pre-trained language model and then fed-forward through a convolutional network. The generated and real poses will then be assessed by a critic network whose objective is to maximize the earth mover's distance between the real and generated samples distributions. Similar to the pose representation in detection-based human pose estimation \cite{tompson2014joint}, we represent the pose by a set of heatmaps each corresponding to a body keypoint. Additionally, to resolve the highly unstable nature of GAN training, we experiment with different GAN models and loss functions and thoroughly evaluate their impact on the synthesized poses. 
	We evaluate the approach on the COCO dataset and show that it is possible to generate human poses that are consistent with a given text.

	\section{Related work}
	Generative models are a powerful tool for learning data distributions. Recent advancements in deep network architectures have enabled modeling complex and high-dimensional data such as images~\cite{salakhutdinov2015learning}. Examples of deep generative models include Deep Belief Networks (DBNs)~\cite{hinton2009deep}, Variational Autoencoder (VAEs)~\cite{kingma2013auto} and the more recent approach of Generative Adversarial Networks (GANs)~\cite{goodfellow14}. In the field of computer vision, GANs have been employed for different tasks for content synthesis, including unconditional image synthesis~\cite{goodfellow14,radford16}, 
	image synthesis conditioned on text~\cite{li2019object,Li_2019_story,NIPS2019_8375,qiao2019mirrorgan,reed16,xu2018attngan,zhang17,tan2019semantics,li2019controllable}, generating text description conditioned on images~\cite{dai2017towards}, style transfer between images~\cite{chu2017cyclegan}, and transferring a target pose to a given person's pose in an image~\cite{ma17}.
	
	Image synthesis conditioned on text has gained traction in computer vision research recently. Motivations for such works include matching features between the semantic and visual space. Reed \etal~\cite{reed16} combine a GAN with a deep symmetric structured text-image joint embedding to synthesize plausible images of birds and flowers from human-written text descriptions. Zhang \etal~\cite{zhang17} propose a GAN composed of two stages and generate hierarchical representations that are transferred between several stacked GANs.
	Reed \etal~\cite{reed16} and Zhang \etal~\cite{zhang17} also attempt to generalize their models to generate images with multiple types of objects using the COCO dataset. However, their approaches do not directly address the human pose, and the persons in the synthesized images have deformed poses. In a subsequent work, Reed \etal~\cite{NIPS2016_6111} generate images based on text and show that using a sparse set of keypoints allows for synthesizing a higher resolution image. Zhou \etal~\cite{zhou2019text} alter the pose of a person in a given image based on a text description. In the first stage they generate a pose by predicting a pose from a set of pose clusters created from the training set. However, they use a pedestrian dataset in which the poses are simple and contain mainly small variations of standing or walking persons. Since the approach assumes that all poses can be represented by a small set of clusters, the approach cannot be applied to datasets with versatile and highly articulated poses such as COCO.
	In a more recent work, Xu \etal~\cite{xu2018attngan} proposed a more advanced attentional GAN, which is multi-stage and attention-based, such that it can synthesize ﬁne-grained details by paying attention to the relevant words in the text. Their model outperforms the previous works but individuals still appear deformed in the generated images. Li \etal~\cite{li2019object} propose an object-driven attention module that generates images conditioned on the class label. However, they do not explicitly handle the human case and the humans still look deformed despite improved results. 
	In the fashion domain, Zhu \etal~\cite{Zhu_2017_ICCV} manipulate the clothing of a person in a given image based on a text description without altering the pose. 
	Other related works such as \cite{grabner2011makes,gupta20113d,Li_2019_affordance,PSI:2019} deal with searching for or synthesizing plausible human poses that match object affordances in a given scene.
	
	\section{Generating human poses from text}
	
	\begin{figure*}[t]
		\begin{center}
			\includegraphics[width=\linewidth]{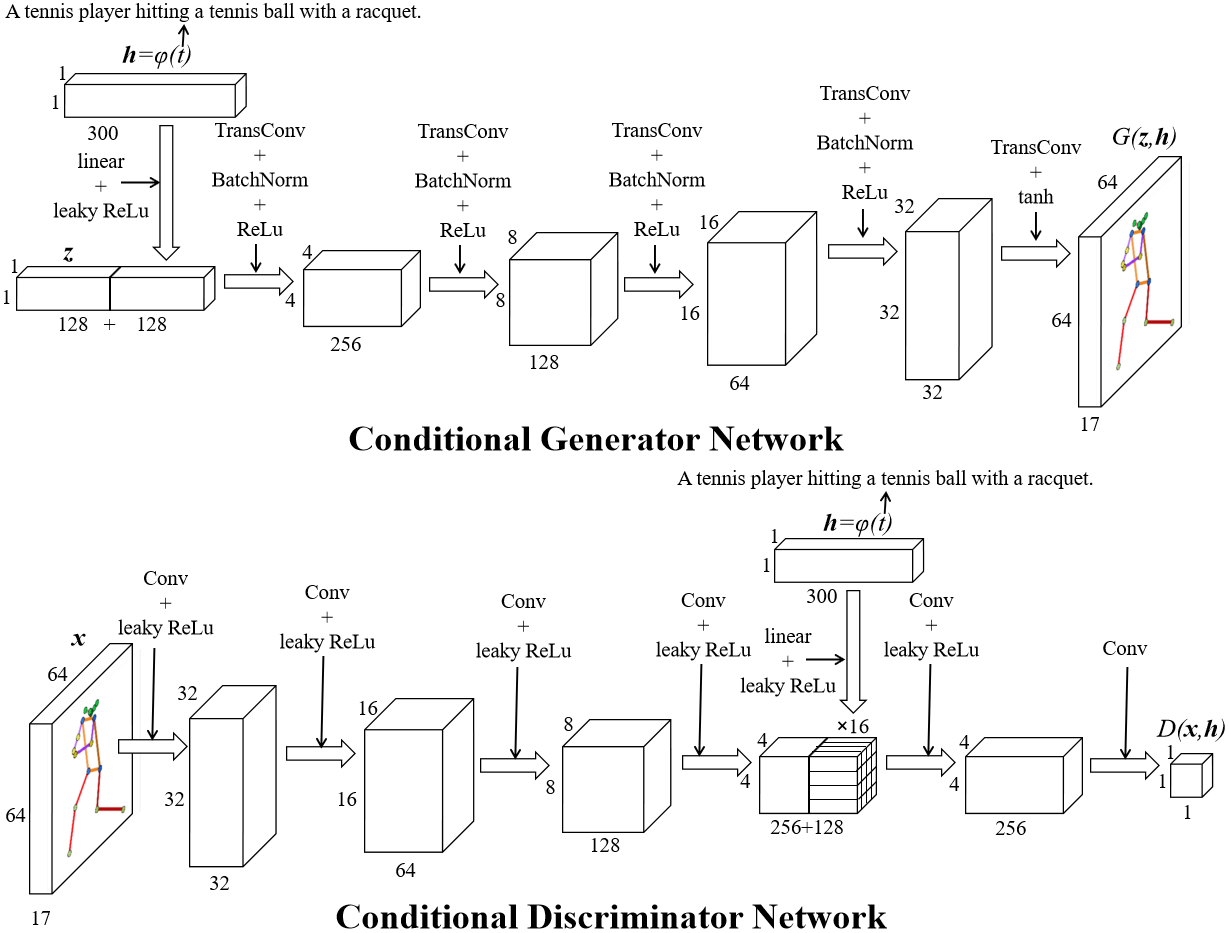}
		\end{center}
		\caption{The architecture of the proposed network. The generator $G$ takes a noise vector $z\in\R^{128}$ and a text encoding vector $h\in\R^{300}$ as input and generates a pose heatmap $G(z,h)\in\R^{J\times 64\times64}$ where J is the number of keypoints. The discriminator $D$ takes a real or generated pose heatmap $H$ and a text encoding vector $h$ as input.  The discriminator predicts a single value $D(x,h)\in\R$ indicating its confidence about the sample being real or generated. For upsampling, transposed convolution layers are used.}
		\label{fig:architecture}
	\end{figure*}
	
	The goal of our approach is to generate human poses that match a textual description as illustrated in Figure~\ref{fig:intro_example}. To this end, we use a conditional Wasserstein GAN as shown in Figure~\ref{fig:architecture}. The text description is first converted into a vector and used to condition the GAN, which predicts heatmaps for each joint, which are finally converted into a human pose. Before we discuss the network architecture in Section \ref{sec:arch}, we discuss the representation of the text and the human pose.
	
	\subsection{Feature representation}\label{sec:feature}
	We need to define representations for the text description as well as the human pose such that they can be used in a convolutional network. The text is encoded by the mapping $\varphi:\mathbb{T}\rightarrow\mathbb{R}^{300}$, which maps a text sequence into a $300$ dimensional embedding space. For the text embedding, we use fastText~\cite{bojanowski2017enriching,mikolov2018advances}. As is common for human pose estimation~\cite{wei16}, we represent the human pose by a heatmap $x \in \R^{m \times n}$ for each joint $j$. The heatmaps are modeled by a Gaussian distribution centered at the keypoint coordinate. Compared to a skeleton representation based on joint coordinates, heatmaps allow for representing joints that are invisible due to occlusion or truncation by setting the heatmaps to zero and allow an implementation based on a convolutional network rather than a fully connected one.  
	The choice of the heatmap-based representation is also validated by our experiments, in which we compare the proposed representation with a skeleton representation that is regressed by a fully connected network. 
	Given these two representations for the text description and the human pose, we will describe the network architecture that generates heatmaps from the embedded text in the following section. 
	
	\subsection{Architecture}\label{sec:arch}
	In order to learn to predict plausible poses from text, we use adversarial training as illustrated in Figure~\ref{fig:architecture}. In our experiments, we show that a vanilla GAN performs poorly. We therefore use a Wasserstein GAN (WGAN), which is a more stable variant for training GANs with their continuous and nearly everywhere differentiable loss functions~\cite{arjovsky17}. 
	
	The model consists of a conditional generator network $G$ and a conditional discriminator network $D$. The input to the generator is a concatenation of a noise vector $z\sim N(0,I)$, where $N$ denotes a normal distribution, with the embedded text description $h=\varphi(t)$ reduced by a layer from $300$ to $128$ dimensions. Given $z$ and $h$, the network infers $J$ heatmaps with resolution $m\times n$, \ie $G(z,h)\in\R^{J\times m\times n}$. The discriminator network takes either real or generated heatmaps as input. Since our goal is to generate heatmaps or poses that match the text description, we condition the network on the embedded text $h=\varphi(t)$ as well. Since the heatmaps have a higher dimensionality with $J\times64\times64$ than $h$, we first apply the inverse transformations of the generator until the resolution is reduced to $4\times4$. We then concatenate the embedded text, by duplicating it $16$ times after a layer that reduces the vector $h$ from $300$ to $128$ dimensions.   
	Both networks are trained together where $D$'s objective is to maximize the distance between the generated heatmaps $G(z,h)$ and the real heatmaps $x$ sampled from the training dataset $\mathbb{P}_r$. Unlike in the unconditional case, $D$ has to distinguish two types of errors: heatmaps that correspond to unrealistic human poses as well as heatmaps that correspond to realistic poses, but the poses do not match the text description. The two errors are penalized by the following two terms:
	
	\begin{equation}
	\begin{aligned}
	L_{D^{*}} = &-\mathbb{E}_{(x,h)\sim\mathbb{P}_r,z\sim\mathbb{P}_{z}}[D(x,h)-D(G(z,h),h)]\\
	& -\mathbb{E}_{(x,h)\sim\mathbb{P}_r,\hat{h}\sim\mathbb{P}_{h}}[D(x,h)-D(x,\hat{h})]
	\end{aligned}
	\end{equation}
	where $(x,h)\sim\mathbb{P}_r$ is a pair of a heatmap and the corresponding text encoding from the training set $\mathbb{P}_r$ and $G(z,h)$ is the generated pose for the same text embedding $h$ and a random noise vector $z$. For the second term, we sample a second text encoding $\hat{h}$ from the training set independently of $x$, \ie $\hat{h}\sim\mathbb{P}_{h}$.
	
	In order to optimize the WGAN using the dual objective of  Kantorovich-Rubinstein~\cite{villani08}, the discriminator network needs to be Lipschitz continuous, \ie $|D(x_2)-D(x_1)|\leq|x_2-x_1|$ for any $x_1, x_2$. Enforcing the Lipschitz constraint requires to constrain the gradient norm of the discriminator to $1$. This can be achieved in two ways. The first approach uses a Lipschitz penality (LP)~\cite{petzka18}:
	\begin{equation}
	R_{LP}= \mathbb{E}_{(\hat{x},h)\sim\mathbb{P}_{\hat{x},h}}[\max(0,\|\nabla_{\hat{x},h}D(\hat{x},h)\|_2-1)^2]. 
	\end{equation}
	The Lipschitz penalty term is one sided and it is only active if the gradient norm is larger than 1. The second approach is termed Gradient Penalty (GP)~\cite{gulrajani17}: 
	\begin{equation}
	R_{GP}=\mathbb{E}_{(\hat{x},h)\sim\mathbb{P}_{\hat{x},h}}[(\|\nabla_{\hat{x},h}D(\hat{x},h)\|_2-1)^2]
	\end{equation}
	which prefers that the gradient is one. In both cases, we sample $\hat{x}$ uniformly along straight lines between a real heatmap $x$ and a generated heatmap $G(z,h)$ conditioned on the matching text encoding $h$, \ie ${\hat{x}=\epsilon x+(1-\epsilon)\cdot G(z,h)}$ where $\epsilon$ is uniformly sampled in $[0,1]$. In our experiments, we evaluate the model when either of these terms is used. The loss function of $D$ is therefore:
	\begin{equation}\label{eq:loss}
	L_D = L_{D^*} + \lambda R
	\end{equation}
	where $R$ is either $R_{LP}$ or $R_{GP}$, which are denoted by WGAN-LP or WGAN-GP, respectively, and $\lambda$ is the regularization parameter for the Lipschitz constraint.  
	To improve the training of $G$, a term with interpolated text encodings is added to the standard loss of $G$:
	\begin{equation}\label{eq:generator_obj}
	\begin{aligned}
	L_G=&-\mathbb{E}_{z\sim\mathbb{P}_{z},h\sim\mathbb{P}_{h}}\left[D(G(z,h),h)\right]\\
	&-\mathbb{E}_{z\sim\mathbb{P}_{z},h_1,h_2\sim\mathbb{P}_{h}}\left[D\left(G\left(z,\frac{h_1+h_2}{2}\right),\frac{h_1+h_2}{2}\right)\right].
	\end{aligned}
	\end{equation}
	Here, $h,h_1,h_2\sim\mathbb{P}_{h}$ are text encodings from the training set, and $\frac{1}{2}h_1+\frac{1}{2}h_2$ is an interpolated encoding between two training samples. The second term adds many more text encoding samples that lie near the real distribution manifold for $G$ to learn~\cite{reed16}.
	
	To obtain poses from the $J$ heatmaps generated by the model, we take the point with the maximum activation in each channel as the location of the corresponding keypoint $j$ if its confidence value is above $0.2$, otherwise we omit the keypoint. This means that our model is not limited to generate full body poses, but it can generate full body poses as well as poses of the upper body only as shown in Figure~\ref{fig:sample1}.

	\section{Dataset and Training}
	
	\paragraph{Dataset.} We use the COCO (Common Objects in Context)~\cite{lin14} dataset for training and evaluating the model. This dataset contains more than $100k$ annotated images of everyday scenes and every image has five human-written text descriptions describing the scene. Additionally, the persons are annotated by $17$ body keypoints. In order to ensure that the text description refers to the person, we only include images which contain a single person and at least $8$ visible keypoints.
	
	\paragraph{Training.}
	We first train an unconditional model, \ie only pose heatmaps are used while the text is excluded. In this way, we pre-train the model on all annotated poses of COCO and we are not limited to the training samples where the text refers to the annotated person, so that the model learns to generate realistic pose.    
	In this setting, the network parameters related to the text encoding are set to zero, while the remaining network parameters are updated.   
	The samples are created by cropping each annotated person using the provided bounding box. In total, there are $116,021$ annotated poses in the training set and $4,812$ poses in the validation set. $G$ is updated after every $5$ iterations of updating $D$. We use $\lambda=10$ as weight for the regularizer in (\ref{eq:loss}).
	
	After pre-training, we train the conditional model using both the pose heatmaps and the text from the images with a single person. For the second stage, there are in total $17,326$ images with a single annotated person in the training set and $714$ images in the validation set.
	During training, we randomly select one of the annotated captions per image. We apply an affine transformation such that the bounding box is located at the center of the image. At this stage, all network parameters are updated and we increase the weight of $\lambda$ to $150$ due to the small number of training samples. 
	To improve training, we also perform some slight data augmentation on the heatmaps by randomly flipping them horizontally and rotating them between $-10^\circ$ and $+10^\circ$ around the center.
	
	\section{Experiments}
	
	\begin{figure*}[t]
		\begin{center}
			\includegraphics[width=\linewidth]{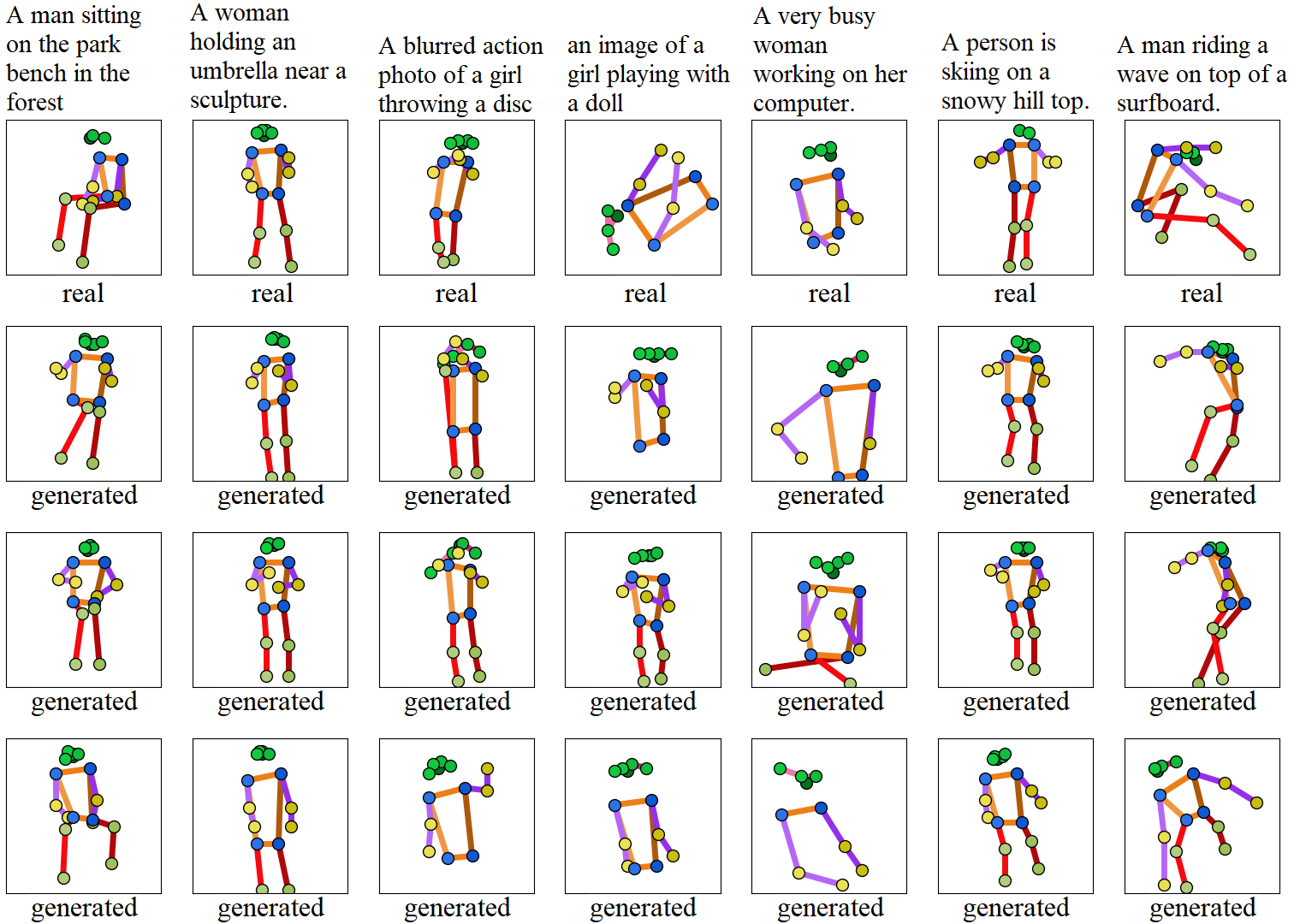}
		\end{center}
		\caption{Examples of generated poses from text. The first row shows the ground-truth pose from the validation set. The text on the top is the associated text. The three poses below each ground-truth pose are synthesized by the model from the text on the top with different noise vectors $z$. It can be seen that some poses such as `throwing' (third column) are more distinct than others such as `holding' (second column). For throwing, we can see that the wrist joint is raised. For `working on the computer' (fifth column), we can see a sitting pose with the wrists extended appearing to be typing.}
		\label{fig:sample1}
	\end{figure*}
	
	\begin{figure*}[t]
		\begin{center}
			\includegraphics[width=1\linewidth]{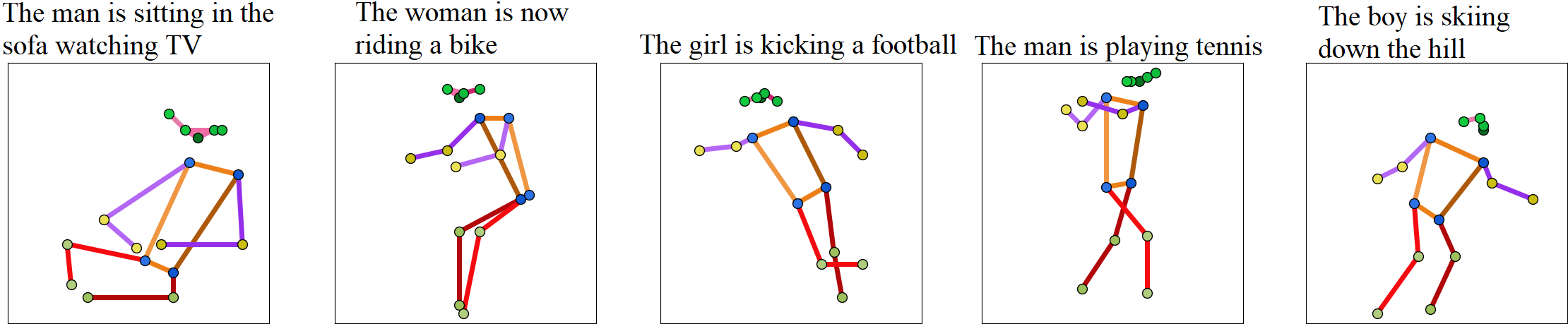}
		\end{center}
		\caption{Poses synthesized from text that is not part of the COCO dataset.}
		\label{fig:customized}
	\end{figure*}
	
	\begin{figure*}[t]
		\begin{center}
			\includegraphics[width=\linewidth]{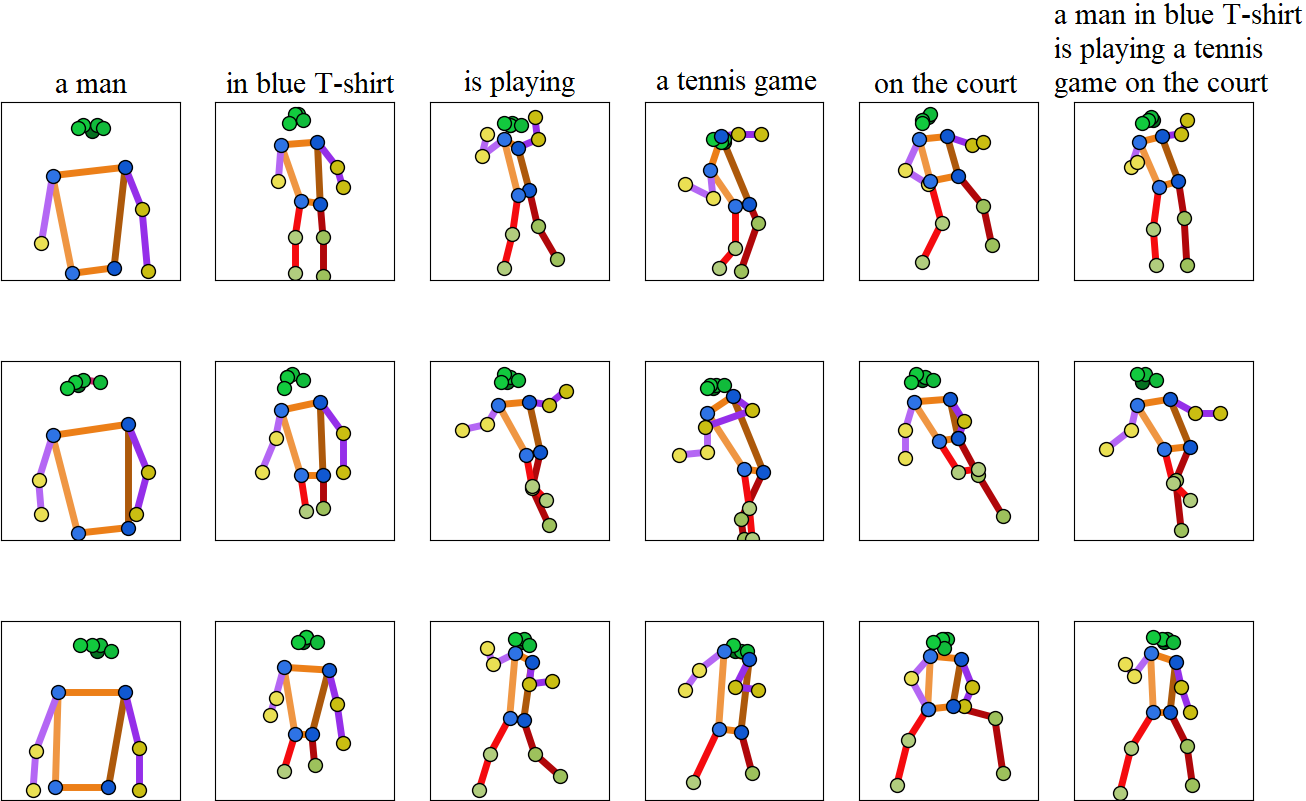}
		\end{center}
		\caption{Poses synthesized from parts of a sentence. The noise input in each row is fixed and varies across the rows. Here what really made the pose unique are the words `tennis game' since the verb `playing' can apply to many different poses.}
		\label{fig:separate}
	\end{figure*}
	
	\paragraph{Qualitative results.} Figure~\ref{fig:sample1} shows some qualitative poses generated by the model, and the ground truth poses as reference. The captions used here are randomly selected from the validation set. We can see that the text encodings are indeed effectively guiding the synthesis of the poses, such that most of the generated poses resemble the real pose and they can reflect the given text, in particular for distinct actions.

	We also evaluated if the model overfits to the text description of the COCO dataset or if it can as well generate plausible poses from other text descriptions as well. Since we do not have any ground-truth poses, we used sentences that relate to activities, such that it is rather clear what the target poses should look like. The results appear in Figure~\ref{fig:customized}. As can be seen, the generated poses match the input text well.
	
	It is also interesting to see what the model can produce if we only feed it with parts of a sentence. Figure~\ref{fig:separate} shows the results. It can be seen that specific verbs and nouns like `playing' and `tennis' matter more in interpreting the context and guiding the model in generating human poses although verbs such as `playing' are generic and can map to various poses, unlike `ski' for example. 
	
	 \begin{figure*}[t]
 		\begin{center}
 			\includegraphics[width=\linewidth]{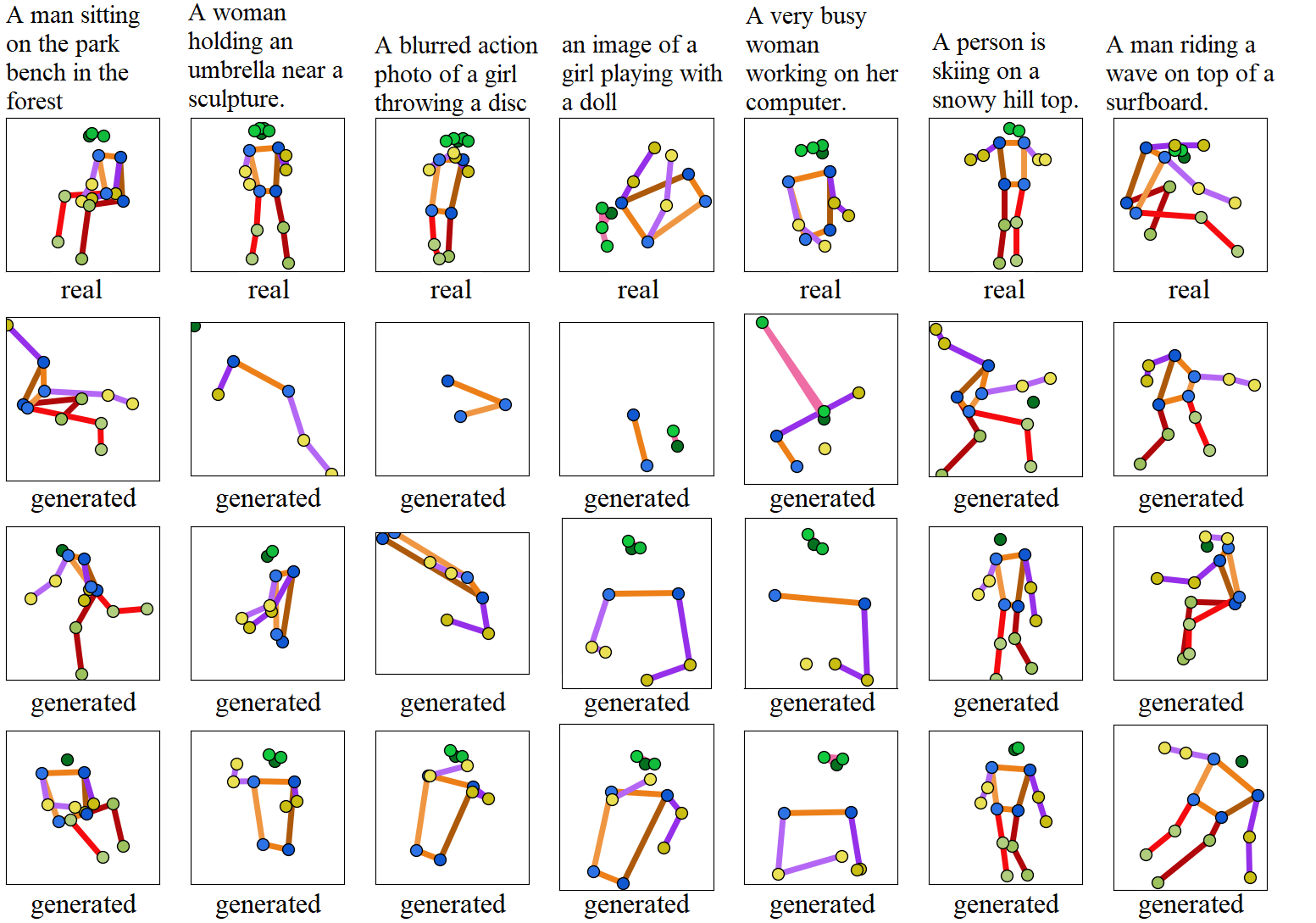}
 		\end{center}
 		\caption{Examples of generated poses using a regression model. 
 		The first row shows the ground-truth pose from the validation set. The text on the top is the associated text. The three poses below each ground-truth pose are synthesized. The synthesized poses of the regression model are clearly worse than the poses synthesized by the proposed model shown in Figure~\ref{fig:sample1}.}
 		\label{fig:sample_regression}
	\end{figure*}
	
	\paragraph{Comparison to regression.} To demonstrate the benefit of representing human poses by heatmaps, we also trained a WGAN-LP that uses a fully connected network to regress the keypoint coordinates instead of a convolutional neural network that predicts a heatmap for each keypoint. In addition to the coordinate prediction, the generator predicts a probability value for the keypoint visibility. For this, we use an additional entropy loss based on the ground truth visibility flags of the training data. The regression approach is less intuitive than the heatmap-based approach and it is more difficult to train.     
	Figure~\ref{fig:sample_regression} shows some qualitative poses generated by the regression model. If we compare the poses with Figure~\ref{fig:sample1}, we clearly see that the regression approach generates less realistic poses than the proposed approach that is based on heatmaps.

	\begin{table*}[t]
		\begin{center}
			\begin{tabular}[h]{l|cccc}
				\hline
				&pose distance &&&\\
				GAN model & $\bar{d}^{p}_{{nn}}$  & $\bar{d}^p_{t_{nn}}$ & 	$\bar{d}^p_{all}$\\ 
				\hline
				Vanilla GAN & $205.2$ & $344.9$ & $351.1$ \\
				WGAN-GP & $82.9$ & $260.5$ & $293.8$ \\
				WGAN-LP Regression & $98.0$ & $268.1$ & $291.1$ \\
				WGAN-LP & $77.2$ & $253.6$ & $287.2$
			\end{tabular}
		\end{center}
		\caption{Quantitative evaluation with respect to the training set. Regression indicates that the keypoint coordinates are regressed instead of being detected using a heatmap representation. 
		}
		\label{tab:quantitaive_evaluation_train} 
		
	\end{table*}

	\paragraph{Quantitative evaluation.} In order to show that the model learned to generate unseen samples that are close to the real distribution, we calculate the distance of the nearest neighbor (NN) pose in the training set of each generated sample conditioned on the text from the validation set and denote it by $\bar{{d}}^{p}_{nn}$. This distance is calculated by generating poses conditioned on the captions from the validation set and then for each such generated pose, we take the distance to its nearest neighbor and finally average the results over all the generated poses. 
	For comparison, in addition to training our algorithm with the Lipshitz-LP term (WGAN-LP), we also train our model using the Lipschitz-GP term (WGAN-GP) and the vanilla GAN. 
	
	Table~\ref{tab:quantitaive_evaluation_train} shows the results.
	The vanilla GAN has the largest distance and we observed that a mode collapse occurs such that there were many repetitions and unrealistic poses in the generated results. 
	When the model is trained using WGAN-GP or WGAN-LP, the NN distance is much smaller where WGAN-LP performs slightly better than WGAN-GP. If a regression-based approach is used instead of the heatmaps, the distance is much higher.    
	The nearest neighbor distance, however, measures only if the generated poses are plausible, but it does not indicate if the generated pose matches the input text. Therefore, in order to show that the text is guiding the pose generation, we calculate the distance to the pose corresponding to the nearest training sample based on the caption, which is obtained by the Euclidean distance in the text embedding space. We denote this distance by $\bar{d}^p_{t_{nn}}$. As for the other distance, WGAN-LP performs slightly better than WGAN-GP and the vanilla GAN performs worst. The regression-based approach performs also worse than the proposed method.     
	We also report the average distance to all poses of the training set, which we denote by $\bar{d}^{p}_{all}$. We provide additional qualitative results for the three approaches in the supplementary material.  

	\begin{table*}[t]
		\begin{center}
			\begin{tabular}[h]{l|ccc|cc}
				\hline
				&pose distance &&& ~~text distance\\
				GAN model & $\bar{d}^p_{nn}$ & $\bar{d}^p_{gt}$ & $\bar{d}^p_{all}$& $\bar{d}^t_{p_{nn}}$ \\  
				\hline
				Vanilla GAN & $218.8$ & $343.2$ & $352.0$ & $10.8$\\
				WGAN-GP & $110.2$ & $255.7$ & $293.4$ & $10.5$\\
				WGAN-LP Regression & $128.2$ & $264.7$ & $290.8$ & $10.5$\\
				WGAN-LP  & $102.3$ & $246.0$ & $286.9$ & $10.5$\\
			\end{tabular}
		\end{center}
		\caption{Quantitative evaluation with respect to the validation set. Regression indicates that the keypoint coordinates are regressed instead of using a heatmap representation. 
		} 
		\label{tab:quantitaive_evaluation_val}
	\end{table*}
	
	To further evaluate the conditional model using the poses in the validation set, we propose the following conditional measure with respect to the validation set. For a text encoding $h_i$ in the validation set, the model synthesizes $k=10$ poses using $k$ different noise vectors $z$. We then calculate three distances for each of the $k$ poses: the first, $\bar{d}^p_{nn}$, is the distance to the nearest neighbor among poses in the validation set; the second, $\bar{d}^{p}_{gt}$, is the distance to the ground truth pose, and the third, $\bar{d}^p_{all}$, is the average distance to all poses in the validation set. Finally, we average the distances over the generated $k$ poses over all samples. The results are reported in Table~\ref{tab:quantitaive_evaluation_val}.
	As for the validation set, we observe that the vanilla GAN struggles to generate realistic poses and WGAN-LP performs slightly better than WGAN-GP. The regression of keypoint coordinates performs also worse than the heatmap representation.
	Furthermore, we calculate the mean distance in the text encoding space. To this end, we obtain for each generated pose the nearest neighbor pose from the validation set. We then compute the distance between the input text and the text of the corresponding nearest neighbor pose. We average the distances over all generated poses. This measure is denoted by $\bar{d}^t_{p_{nn}}$. The differences are smaller compared to the pose distances, but it still shows that the WGANs outperform the vanilla GAN.

	\begin{figure*}[t]
		\begin{center}
			\includegraphics[width=\linewidth]{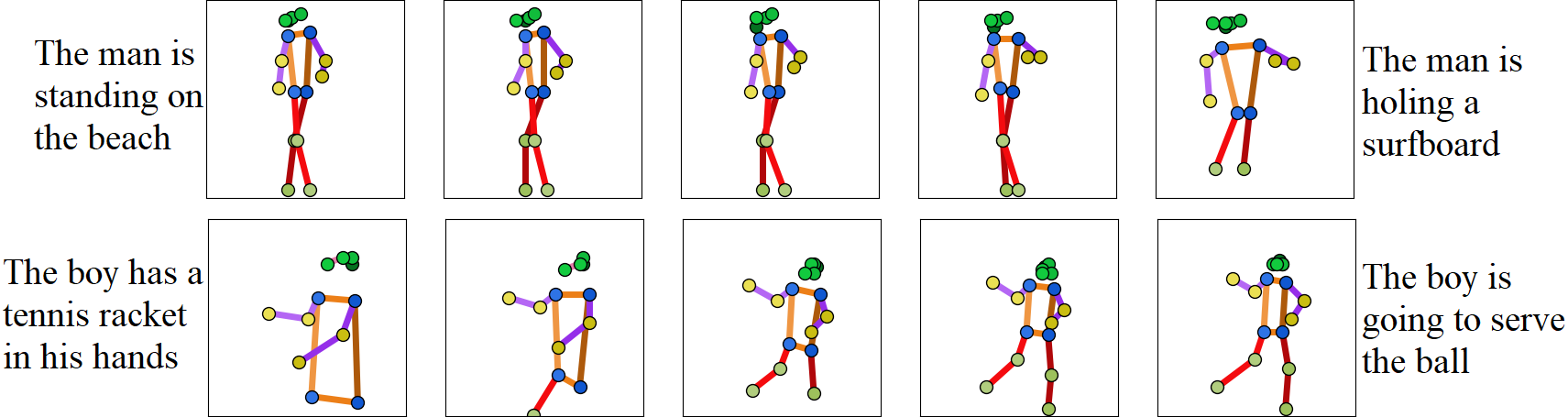}
		\end{center}
		\caption{Interpolation results of text encoding. In each row, the leftmost and rightmost poses are synthesized from the captions. The three poses in the middle are synthesized from interpolations of the encodings of the two captions while $z$ is kept fixed.}
		\label{fig:interpolation}
	\end{figure*}
	
	\paragraph{Interpolation test.} Another interesting qualitative measure is the interpolation between two text descriptions and observing the generated poses. If the generated poses show smooth transitions between the interpolations, we can conclude that the model learned a proper distribution instead of just having memorized the training samples~\cite{borji19}. Given two embedded text descriptions $h_1$ and $h_2$, we interpolate between them by $\hat{h} = w\cdot h_1 + (1-w)\cdot h_2$ with $w\in\{1, 0.75, 0.5, 0.25, 0\}$. For this experiment, we keep the noise $z$ fixed. Figure~\ref{fig:interpolation} shows two interpolation examples. In the first example, we interpolate between `The man is standing on the beach' and `The man is holding a surfboard'. We observe that the right arm gradually moves up for the holding pose. We also observe that the full body pose is generated at the beginning, but the camera gets closer on the right hand side and only two-thirds of the person are visible. The second example interpolates between `The boy has a tennis racket in his hands' and `The boy is going to serve the ball'.    
	
		\begin{table*}[t]
		\begin{center}
			\begin{tabular}[h]{ccc}
				\hline
				Real pose & Generated pose & Equally well\\
				\hline
				$48.81\%$ & $35.31\%$ & $15.88\%$ \\
			\end{tabular}
		\end{center}
		\caption{The percentage of the users choosing the matching pose as the real pose, generated pose or ``equally well''.} \label{tab:subjective_evaluation}
	\end{table*}
	
	\paragraph{User study.} For the subjective evaluation, we have designed an online questionnaire in which $20$ text descriptions from the validation set are taken. For each text description, a user is presented with two human poses, in which one is the real pose matching the text, and the other is synthesized by the model conditioned on this text. The $20$ captions are randomly selected from the validation set and the generated poses have not been cherry-picked. The user is asked to choose which of the two poses matches the caption better or if they match the text equally well. The results are summarized in Table~\ref{tab:subjective_evaluation}. Eighty people in total participated in the survey. The ratio between choosing generated and real poses is around 5:7. And for more than $50\%$ of the time, the users cannot correctly distinguish the generated pose from the real one, \ie, they either choose the generated pose or rate the poses equally well.

% 		\begin{figure*}[t]
% 		\begin{center}
% 			\includegraphics[width=0.8\linewidth]{images/sample_regression.png}
% 		\end{center}
% 		\caption{Some sample outputs of the regression model trained with WGAN-LP.}
% 		\label{fig:sample_regression}
% 	\end{figure*}

	\section{Conclusion}
	
	In this work, we have addressed the task of human pose synthesis from text for highly complex poses. We have designed an effective model using a conditional Wasserstein GAN that generates plausible matching poses from text descriptions. We have demonstrated by qualitative and quantitative results on the COCO dataset that the proposed approach is effective, and additionally outperforms a vanilla GAN and a regression-based approach. We have also conducted a user study that confirmed our results. The model was also able to interpolate poses between two text descriptions. Furthermore, we have shown that the model generalizes well and can additionally generate plausible poses for unseen sentences that are not part of the COCO dataset. 
	\paragraph{Acknowledgement.} The work has been funded by the Deutsche Forschungsgemeinschaft (DFG, German Research Foundation) GA 1927/5-1 and the ERC Starting Grant ARCA (677650). 
	\bibliographystyle{splncs04}
	\bibliography{egbib}
	
\end{document}

\pagestyle{headings}
	\mainmatter

	% Insert your submission number here
	\def\GCPR20SubNumber{45}

	% Replace with your title
	\title{Adversarial Synthesis of Human Pose From Text}

	% DO NOT MODIFY these for the draft version that is used for the
	% review process.
	\titlerunning{Adversarial Synthesis of Human Pose From Text}
	\authorrunning{Zhang, Briq, Tanke, Gall}
	\author{Yifei Zhang\textsuperscript{1,2}
	\and
	Rania Briq\textsuperscript{1}
	\and
	Julian Tanke\textsuperscript{1}
	\and
	Juergen Gall\textsuperscript{1}}
	\institute{Computer Vision Group, University of Bonn\\
	{\tt\small briq,tanke,gall@iai.uni-bonn.de}
	\and
	Bonn-Aachen International Center for Information Technology, RWTH-Aachen University\\
	{\tt\small yifei.zhang@rwth-aachen.de}
	}
	
	\maketitle
	\section{Supplementary}
\begin{figure}[t]
		\begin{center}
			\includegraphics[width=1\linewidth]{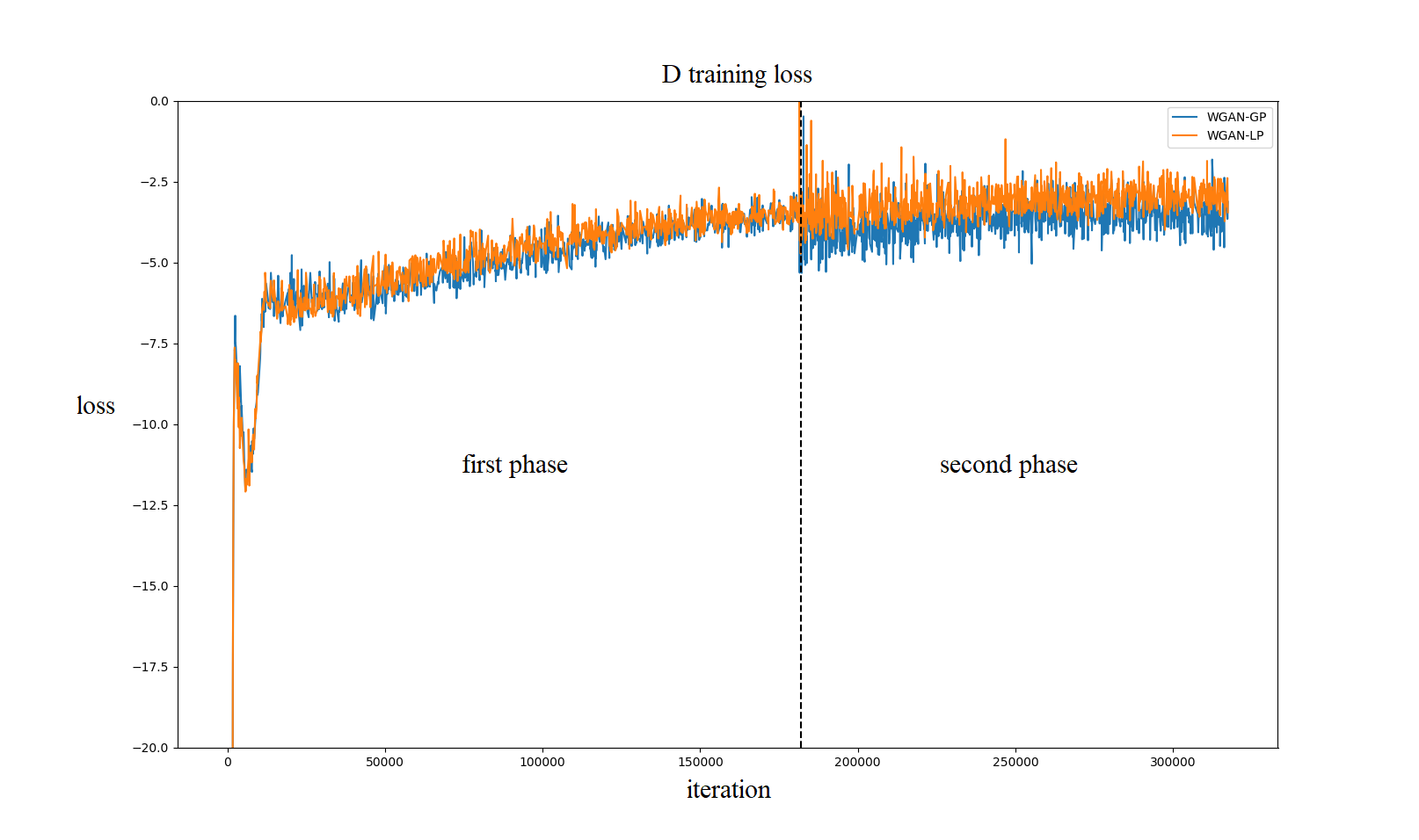}
			\includegraphics[width=1\linewidth]{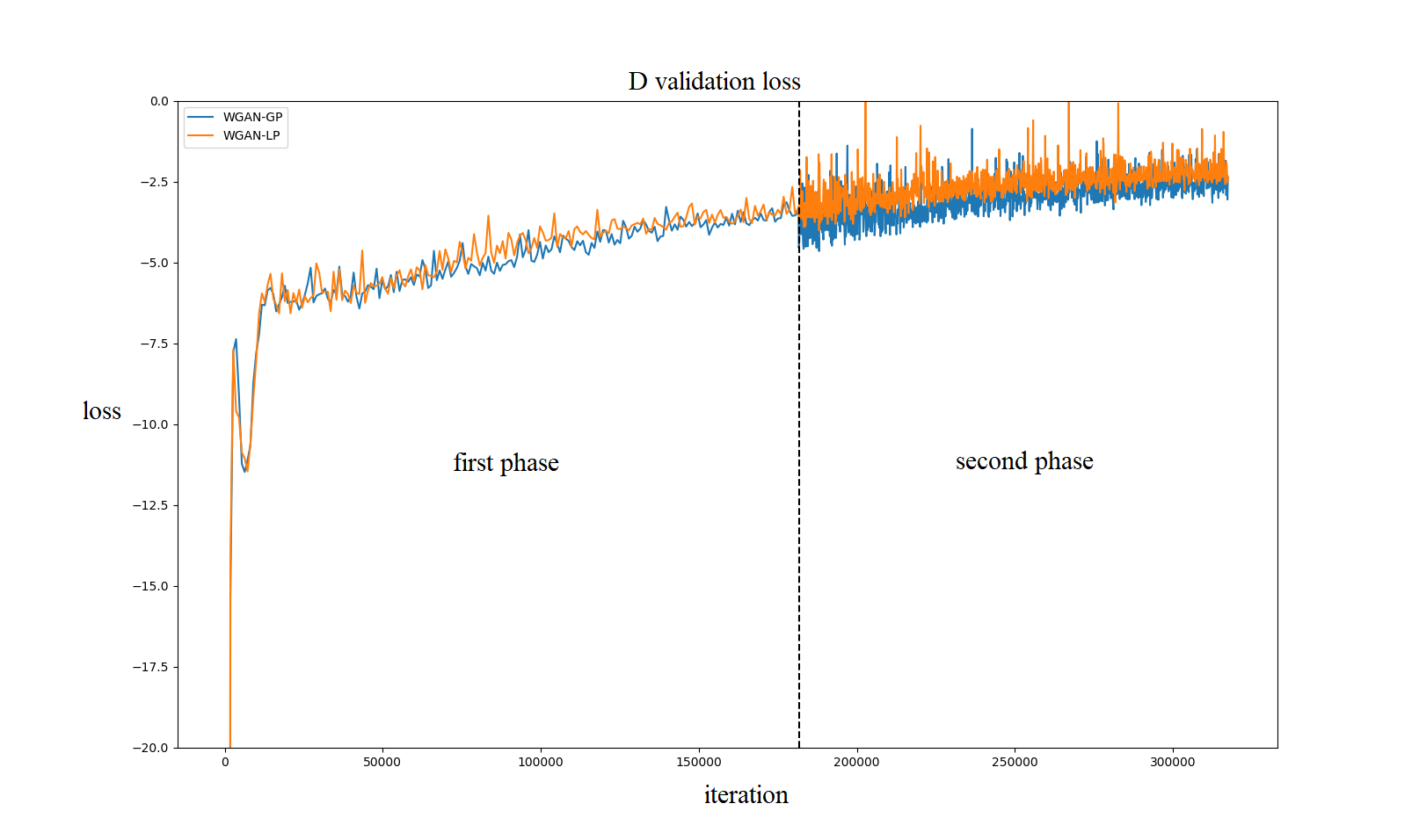}
		\end{center}
		\caption{The training and validation loss curves of the critic $D$ during the two training phases for the two WGAN variants. The orange and blue curve correspond to WGAN-LP and WGAN-GP, respectively.}
		\label{fig:curve}
	\end{figure}

\paragraph{Loss Curves.}	In WGANs, the loss is known to be an indication of the quality of the generated samples where its value indicates the distance to the true distribution.
	In Fig.~\ref{fig:curve}, we show the loss curves of the generators of WGAN-GP and WGAN-LP. We can observe that the loss of two WGAN models decreases (in the absolute value) across the training iterations, indicating that the generator is learning to generate plausible poses and is improving over time. However, we observe that the loss curve of WGAN-GP decreases slightly less than WGAN-LP and more slowly, especially in the second training phase where $\lambda=150$ compared to $10$ in the first phase, and such a large value has been shown to deteriorate training substantially [19], although in our results the deterioration is not substantial.
	
\begin{figure*}[t]
		\begin{center}
			\includegraphics[width=0.8\linewidth]{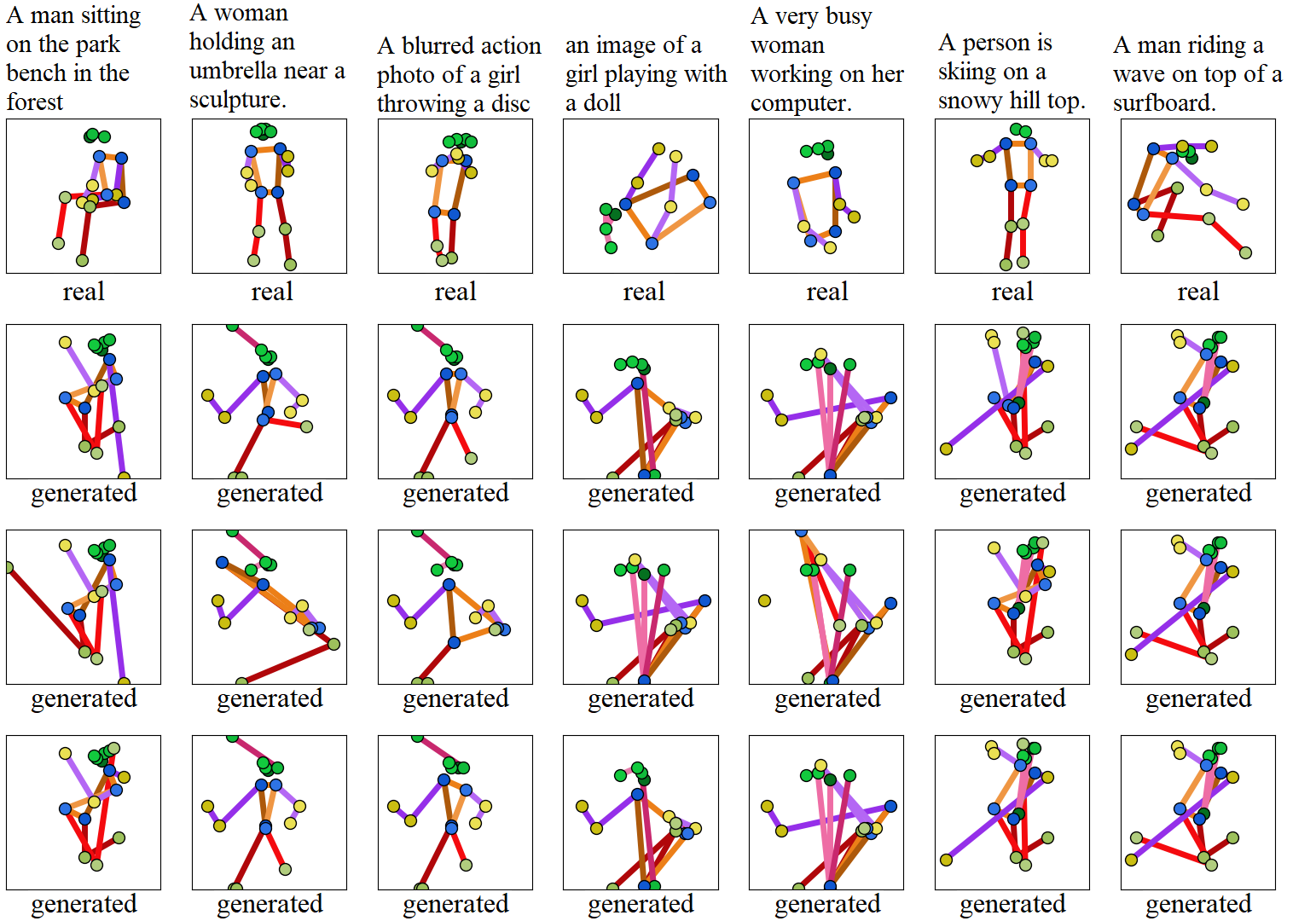}
		\end{center}
		\caption{Some sample outputs of the model trained with the Vanilla GAN. The first row is the ground-truth from the validation set. The text on the top is the associated text. The three poses below each real pose are synthesized by the model from the text on the top with different noise vectors $z$.}
		\label{fig:sample_gan}
	\end{figure*}

\begin{figure*}[t]
		\begin{center}
			\includegraphics[width=0.8\linewidth]{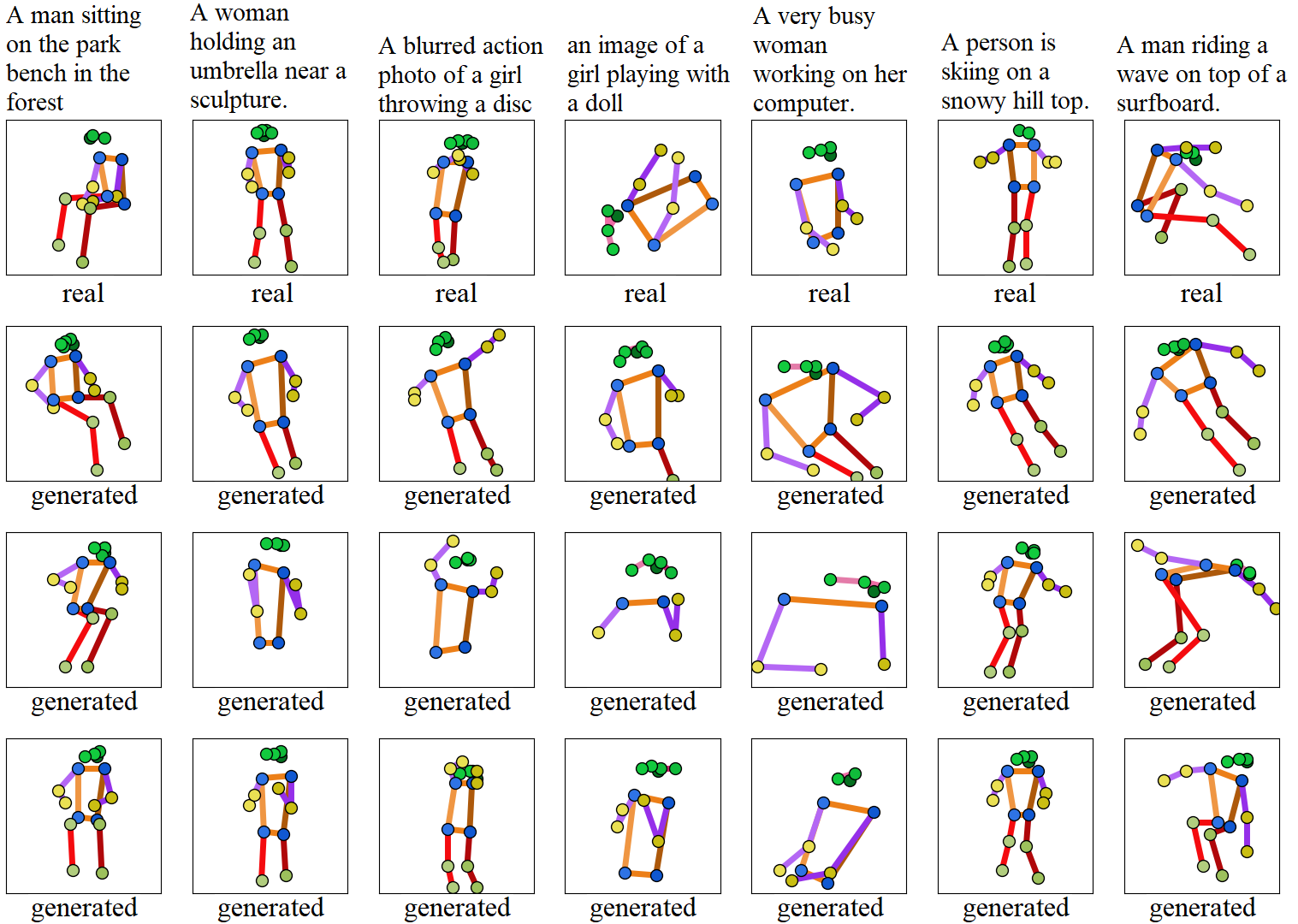}
		\end{center}
		\caption{Some sample outputs of the model trained with the GP term (WGAN-GP).}
		\label{fig:sample_gp}
	\end{figure*}

\paragraph{Qualitative Results.} 	We also include additional qualitative results to point out the differences in the synthesized poses stemming from changing the underlying GAN model. Fig.~\ref{fig:sample_gan} shows the synthesized poses of Vanilla GAN. While sometimes the poses look realistic and consistent  with the input text, changing the noise vector resulted often in very unrealistic poses due to mode collapse. In both WGAN variants, the results look much better than the vanilla GAN. \\
Fig.~\ref{fig:sample_gp} corresponds to the WGAN-GP. 

\begin{figure*}[t]
		\begin{center}
			\includegraphics[width=0.8\linewidth]{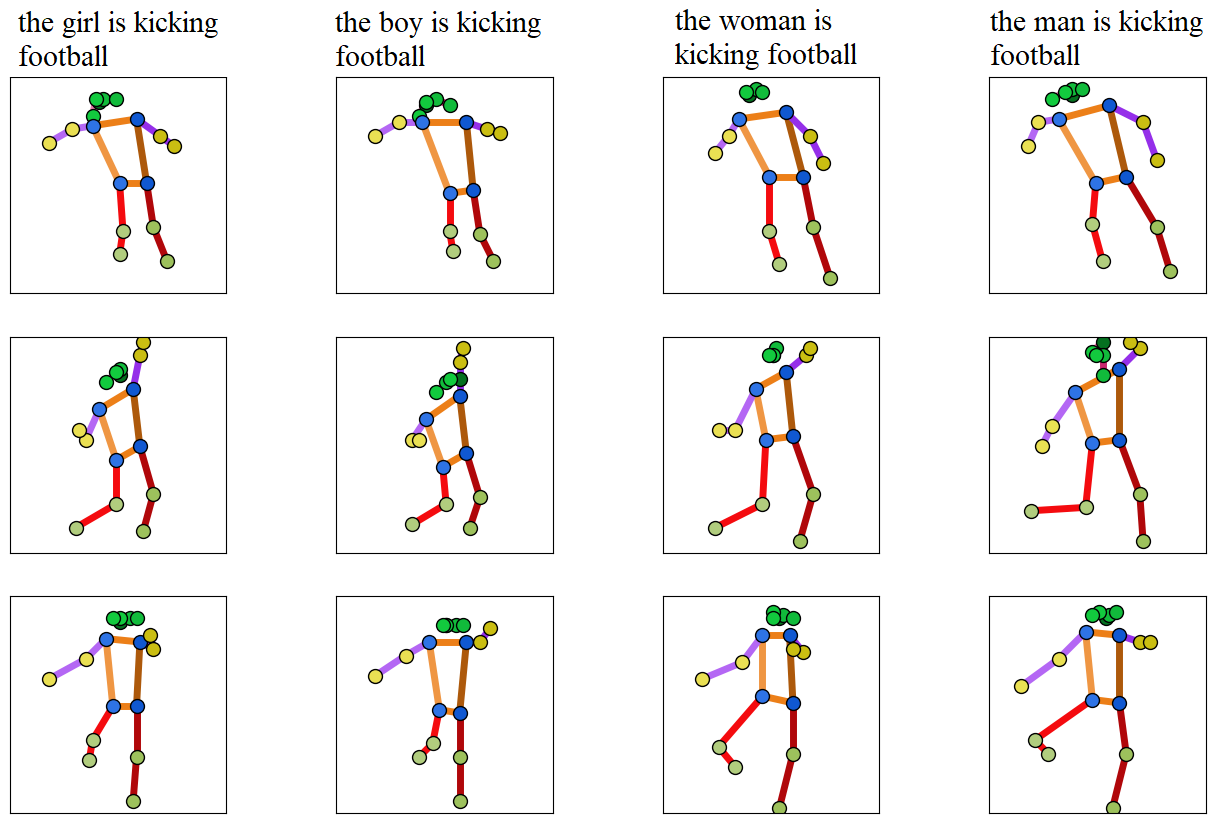}
		\end{center}
		\caption{Poses synthesized from captions with different subject genders and age. The caption to synthesize each column of poses is on the top. The noise input is the same for each row.}
		\label{fig:gender}
	\end{figure*}

\paragraph{Subject Pose.} We are also interested in how the generated pose changes based on the subject, e.g. an adult versus a young person. Fig.~\ref{fig:gender} shows some generated poses for the same caption with different subjects while keeping the noise fixed. The generated poses show subtle difference between the different genders. But if we look more closely, we can find that a young person's pose is slightly smaller than an adult's pose, which reflects the reality. This indicates that for pose generation, the subject of the caption does not matter very much, and what really matters is the action.

\begin{figure*}[t]
		\begin{center}
			\includegraphics[scale=0.8]{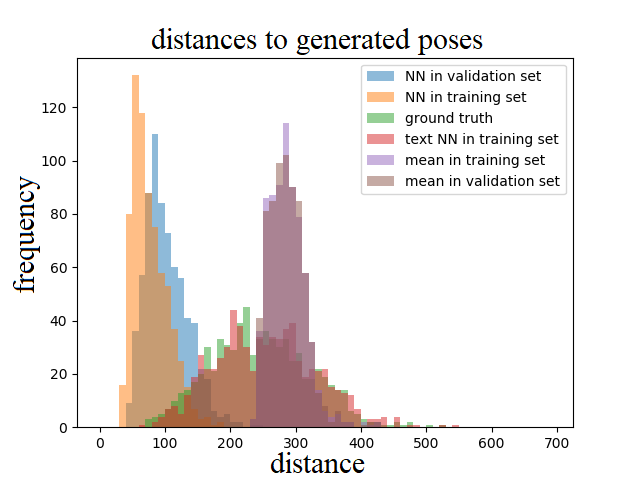}
		\end{center}
		\caption{WGAN-LP. Histograms of pose distances.}
		\label{fig:hist}
	\end{figure*}

\begin{figure*}[t]
		\begin{center}
			\includegraphics[scale=0.8]{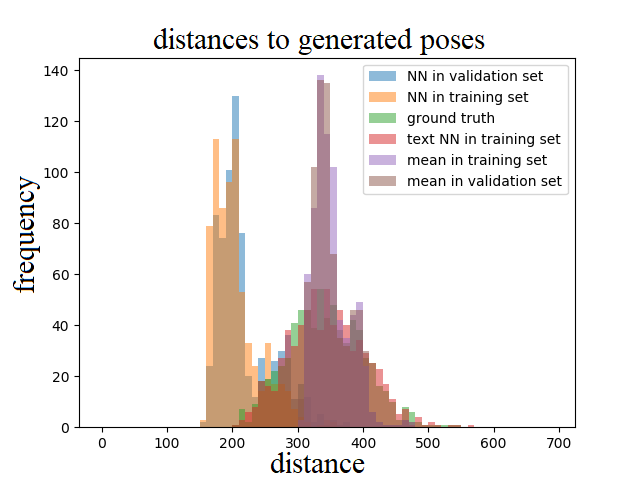}
		\end{center}
		\caption{Vanilla GAN. Histograms of pose distances.
		}
		\label{fig:hist_gan}
	\end{figure*}

\begin{figure*}[t]
		\begin{center}
			\includegraphics[scale=0.8]{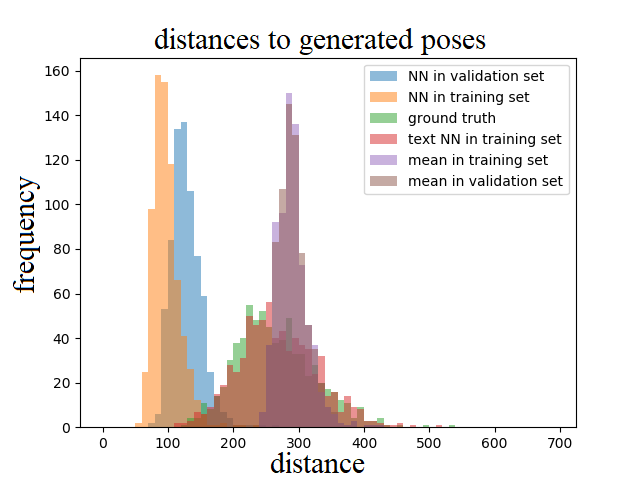}
		\end{center}
		\caption{WGAN-LP Regression. Histograms of pose distances.
		}
		\label{fig:hist_regression}
	\end{figure*}

\paragraph{Quantitative Results.} In Fig.~\ref{fig:hist}, \ref{fig:hist_regression} and \ref{fig:hist_gan}, we plot the pose distance histograms corresponding to table 1,2 from the manuscript for the WGAN-LP, regression and vanilla GAN and WGAN-LP to show the distribution of distances. In Fig.~\ref{fig:hist}, we can see that the generated poses' distances to their NN poses (blue and orange) are much smaller than their distances to all poses on average (purple and brown), while their distances to their ground truth (green) and text-NN (red) poses are shifted away from the average distances towards the NN distances, meaning that in the model the text encodings are indeed guiding the poses synthesis toward the correct direction. For the regression (Fig.~\ref{fig:hist_regression}), such phenomenon is less evident. For the Vanilla GAN (Fig.~\ref{fig:hist_gan}), such phenomenon is even much less evident.

\begin{figure}[t]
		\begin{center}
			\includegraphics[width=1\linewidth]{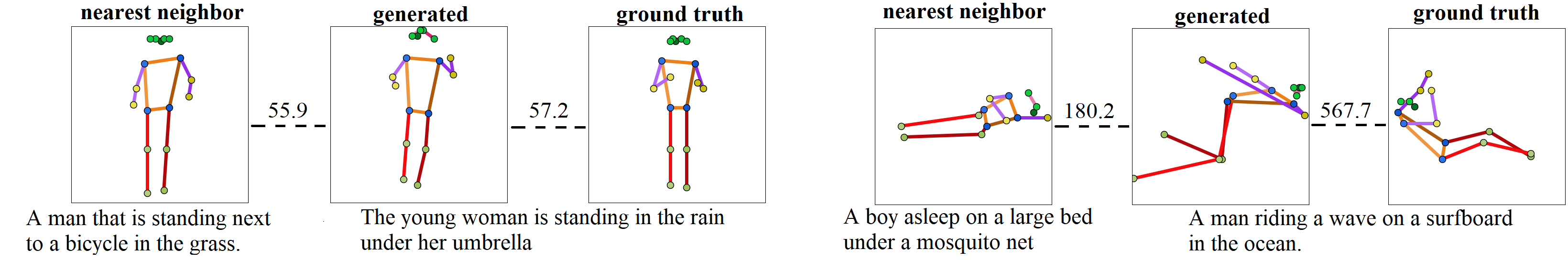}
		\end{center}
		\caption{Two generated poses, their ground truth poses, and their nearest neighbor poses in the validation set. The text descriptions are below the poses and the distances are shown between them. Left: the ground truth is close to the generated pose and the nearest neighbor has a similar text description. Right: the ground truth is far from the generated pose and the nearest neighbor has a very different text description. However, the large distance to the ground truth is due to the opposite orientation of the pose.}
		\label{fig:far_near}
	\end{figure}

In Fig.~\ref{fig:far_near}, we show why a generated pose is sometimes far from the ground truth pose, even though it looks plausible for the given input text. 
	
\begin{figure*}[t]
		\begin{center}
			\includegraphics[width=\linewidth]{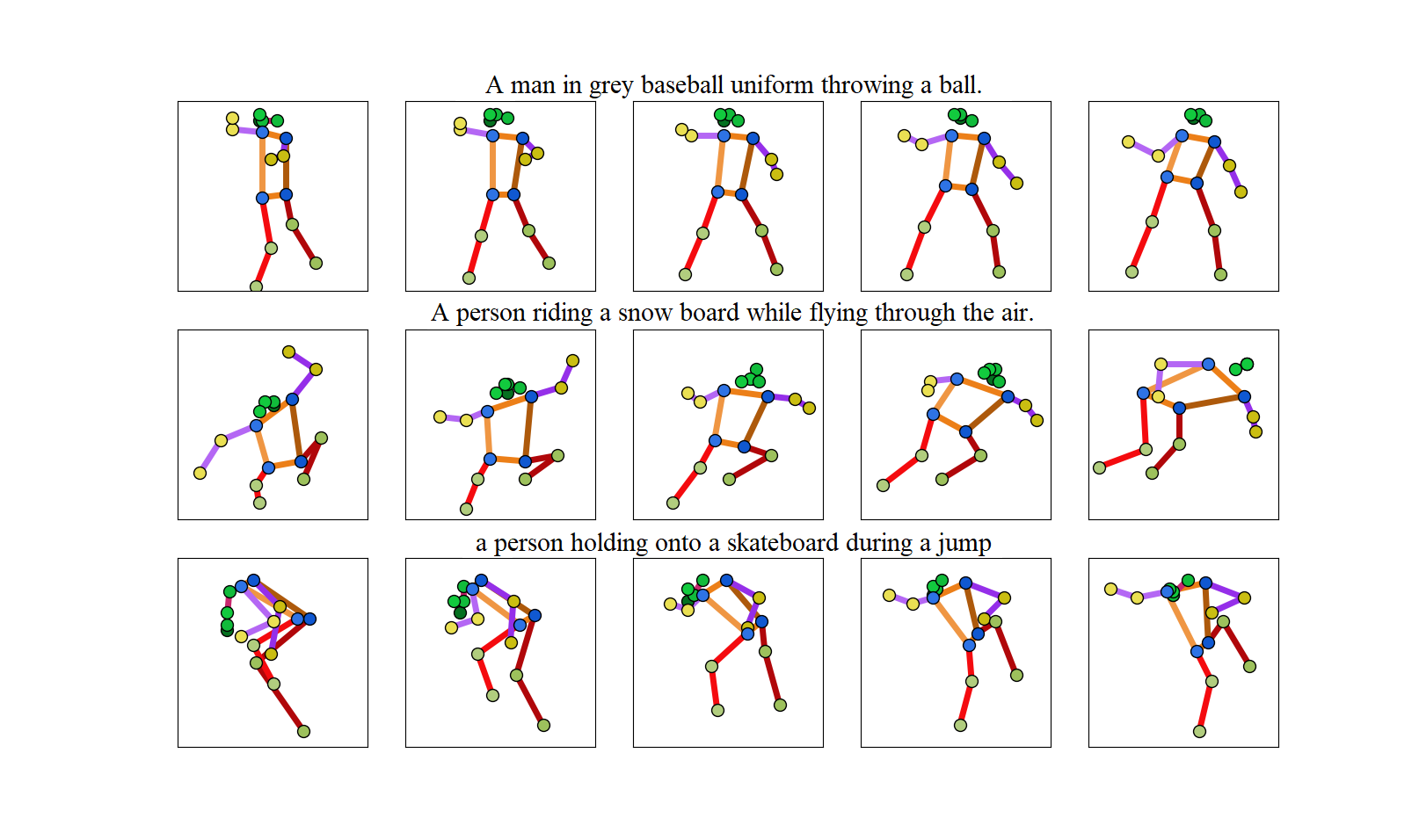}
		\end{center}
		\caption{Interpolation results of noise input. In each row, the five poses are synthesized from the text on the top. The noise inputs of the three poses in the middle are interpolated between the noise inputs of the leftmost and rightmost poses.}
		\label{fig:interpolation_noise}
	\end{figure*}

\paragraph{Noise Interpolation Test.}	Similarly to the text interpolation test, we also perform a noise interpolation test, where the text is kept fixed and the noise vector is interpolated. As in the text interpolation test, we observe smooth transitions over the interpolated noise vector in Fig.~\ref{fig:interpolation_noise}.
	
